%% file: main.tex
\documentclass[11pt]{article} 

\usepackage[margin=1in]{geometry}
\usepackage{times}

\input{math_commands.tex}

\usepackage{hyperref}
\usepackage{url}
\usepackage{graphicx}
\usepackage{xcolor}
\usepackage{booktabs}
\usepackage{pgfplots}
\pgfplotsset{compat=1.18}
\usepackage{tikz}
\usepackage{algorithm}
\usepackage{algpseudocode}
\usetikzlibrary{arrows,positioning,calc,fit,shapes.geometric,backgrounds}
\usepackage{tabularx}
\usepackage{array}
\usepackage[numbers,sort&compress]{natbib}
\newcolumntype{Y}{>{\raggedright\arraybackslash}X}

\title{Mixture of Chapters: Scaling Learnt Memory in Transformers}

\author{%
\parbox{0.9\textwidth}{\centering
{\large
Tasmay Pankaj Tibrewal$^{1,2}$ \qquad
Pritish Saha$^{1}$ \qquad
Ankit Meda$^{1}$ \\
[0.3em]
Kunal Singh$^{2}$ \qquad
Pradeep Moturi$^{2}$
}\\[0.5em]
{\normalsize
$^{1}$IIT Kharagpur \qquad $^{2}$Fractal AI
}\\[0.25em]
{\ttfamily\small
tasmay.tibrewal@kgpian.iitkgp.ac.in,
pritish.saha@kgpian.iitkgp.ac.in,\\
ankitm18@kgpian.iitkgp.ac.in,
kunal.singh@fractal.ai,
pradeep.moturi@fractal.ai
}
}}

\begin{document}
\maketitle

\begin{abstract}
Transformers lack an explicit architectural mechanism for storing and organizing knowledge acquired during training. We introduce learnable sparse memory banks: a set of latent tokens, randomly initialized and trained end-to-end, that transformer layers query via cross-attention to retrieve stored knowledge. To scale memory capacity without prohibitive attention costs, we propose chapter-based routing inspired by Mixture-of-Experts architectures, partitioning the memory bank into chapters and training a router to select relevant subsets per input. This enables scaling to 262K memory tokens while maintaining tractable computation. We evaluate our approach against standard transformers (in iso-FLOP settings) on pre-training and instruction fine-tuning across relevant benchmarks. Our models surpass iso-FLOP baselines suggesting scope for a new axis of scaling, demonstrating that explicit associative memory provides complementary capacity to what is captured implicitly in model parameters. Additionally, we observe improved knowledge retention under continued training, with robustness to forgetting when transitioning between training phases (e.g., pretraining to instruction fine-tuning). 
\end{abstract}

\section{Introduction}
Transformers provide a powerful foundation for sequence modeling \citep{vaswani2017attention}, but they do not expose an explicit architectural mechanism for persistent, addressable memory. In practice, many systems instead rely on inference-time mechanisms such as segment-level recurrence over prior hidden states \citep{dai2019transformerxl}, KV-cache retention, eviction, or compression methods for long-context generation \citep{xiao2023streamingllm,zhang2023h2o,chen2024nacl,karami2025trellis}, tool-based updates \citep{schick2023toolformer,yao2023react}, and retrieval-augmented generation (RAG) \citep{lewis2020retrieval}. These methods are effective, but they primarily reuse or manage previously processed context, or retrieve external information, rather than providing a learned internal memory store for factual knowledge acquired during training.

We study a complementary direction: a transformer augmented with a learned memory bank \citep{wu2020memformer,sukhbaatar2019memory}. The memory is a set of latent tokens, randomly initialized and trained end-to-end, that can be queried via cross-attention, where the token stream produces queries and the memory provides keys and values \citep{wu2020memformer}. This design stores knowledge in a compact latent space rather than as retrieved text, and it can be integrated inside standard transformer blocks \citep{wu2020memformer,jaegle2021perceiverio}.

The main bottleneck is scale. Dense cross-attention over a very large bank is expensive. We therefore introduce \textbf{Mixture of Chapters (MoC)}\footnote{Code is available at \url{https://github.com/Tasmay-Tibrewal/Memory}.}, which partitions the memory bank into chapters and uses a lightweight router to select a small subset of chapters per input at the sequence level \citep{shazeer2017moe,fedus2022switch,roy2021routing}. This enables scaling learned memory capacity to large sizes while keeping computation tractable, and it fits naturally with the goals of associative memory research \citep{berges2025memorylayers,lample2019productkey}.

\section{Motivation}
We motivate \textbf{Mixture of Chapters} with three goals.

\paragraph{(1) Explicit memory.}
Transformers store knowledge implicitly in dense parameters \citep{vaswani2017attention}, making it hard to inspect, edit, or scale memory independently. Prior work adds explicit memory via large key-value tables \citep{lample2019pkm,berges2024memorylayers} or attention-based memory modules \citep{wu2020memformer,dejong2021mentionmemory}. We follow this direction by introducing a learned, addressable memory bank that complements parametric capacity.

\paragraph{(2) Scalable sparse access.}
Dense attention over large memory banks is expensive \citep{lample2019pkm}. Inspired by sparse routing in MoE models \citep{shazeer2017moe,fedus2021switch} and routing-based sparsification \citep{roy2020routing}, we partition memory into chapters and route each input to a small subset, enabling predictable scaling at tractable cost.

\paragraph{(3) Retention under continued training.}
Post-training shifts can cause catastrophic forgetting \citep{kirkpatrick2017ewc}, including in LLM instruction tuning \citep{luo2023cfllm}. An explicit memory bank can reduce interference during continued training by anchoring factual content, which we observe empirically across training phase transitions \citep{cossu2022continualpretraining}.

\section{Related Work}
\paragraph{Most relevant.}
Our work is closest to learned, scalable internal memory modules integrated into transformers. Product Key Memory (PKM) provides a classic recipe for large trainable key-value memory with efficient sparse lookup \citep{lample2019pkm}. Memory Layers at Scale shows that trainable memory layers can add substantial capacity with near-constant FLOPs and strong factual gains \citep{berges2025memorylayers}. We build on this direction, but use a learned latent-token memory bank accessed via cross-attention and scale it with sequence-level chapter routing.

\paragraph{Other memory mechanisms.}
A broad set of alternatives externalize memory at inference time or focus on efficiency and long-context handling. Transformer-XL exemplifies recurrent caching to extend effective context \citep{dai2019txl}, and RAG exemplifies retrieval over an external corpus \citep{lewis2020rag}. Conditional computation via routing, as in mixture-of-experts, provides the sparse-activation template we adapt for memory selection \citep{shazeer2017moe}. Finally, continual training can induce catastrophic forgetting \citep{kirkpatrick2017ewc}, which motivates our retention experiments. A wider map of related directions is provided in Appendix~\ref{sec:litreview} (Table~\ref{tab:lit_map}).

\section{Contributions}
This paper makes the following contributions:
\begin{itemize}
    \item \textbf{Learned associative memory bank in transformers.} We add a persistent bank of latent memory tokens trained end-to-end and accessed via cross-attention, enabling internal retrieval in a compact latent space rather than text-based retrieval \citep{wu2020memformer,berges2024memorylayers}.
    \item \textbf{Mixture of Chapters for scalable sparse access.} We partition memory into chapters and train a lightweight router that selects a small subset per input, adopting the conditional computation principle from Mixture of Experts \citep{shazeer2017moe,fedus2021switch}.
    \item \textbf{Scaling behavior under fixed compute.} We evaluate our memory architecture against standard transformers in iso-FLOP settings, complementing compute-optimal scaling discussions in language modeling \citep{hoffmann2022chinchilla}.
    \item \textbf{Retention under continued training.} We provide evidence that explicit memory improves knowledge retention and reduces forgetting across training phase transitions, connecting to continual learning analyses in LLMs \citep{luo2023cfllm,kirkpatrick2017ewc}.
\end{itemize}

\begin{figure*}[t]
\centering
\begin{tikzpicture}[
  >=latex,
  font=\scriptsize,
  node distance=2.8mm and 16mm
]
\definecolor{cNorm}{RGB}{232,224,255}
\definecolor{cSA}{RGB}{232,224,255}
\definecolor{cMem}{RGB}{255,220,175}
\definecolor{cMLP}{RGB}{195,255,205}
\definecolor{cRouteBG}{RGB}{232,255,235}
\definecolor{cRouteBox}{RGB}{255,205,205}
\definecolor{cRouteDia}{RGB}{255,230,180}
\definecolor{cBank}{RGB}{210,255,215}

\tikzset{
  blk/.style={draw, rounded corners=2pt, thick, align=center, inner sep=2.5pt, minimum width=44mm, minimum height=6.5mm},
  plus/.style={draw, circle, thick, inner sep=0pt, minimum size=4.6mm, font=\scriptsize\bfseries},
  note/.style={font=\scriptsize, align=left},
  cont/.style={draw, rounded corners=4pt, thick, inner sep=5pt},
  arrow/.style={->, thick},
  resid/.style={->, thick, dashed, color=orange!80!black, rounded corners=2pt},
  kv/.style={->, thick, color=red!70!black},
  routearrow/.style={->, thick, color=orange!80!black},
  routedia/.style={draw, shape=diamond, thick, aspect=1.8, inner sep=1.2pt, align=center, fill=cRouteDia},
  chap/.style={draw, rounded corners=2pt, thick, inner sep=1.5pt, minimum width=9mm, minimum height=5mm, align=center},
  bankhdr/.style={font=\scriptsize, align=center},
  lab/.style={font=\scriptsize, inner sep=1pt, fill=white, text opacity=1, fill opacity=0.9}
}

\node[blk, fill=cNorm] (n1) {RMSNorm};
\node[blk, fill=cSA, below=of n1] (sa) {\textbf{Self-attn}\\{\scriptsize 12H, 4KV (GQA); RoPE}};
\node[plus, below=of sa] (add1) {+};

\node[blk, fill=cNorm, below=of add1] (n2) {RMSNorm};
\node[blk, fill=cMem, below=of n2] (mem) {\textbf{Mem cross-attn}\\{\scriptsize 12H, 12KV}};
\node[plus, below=of mem] (add2) {+};

\node[blk, fill=cNorm, below=of add2] (n3) {RMSNorm};
\node[blk, fill=cMLP, below=of n3] (mlp) {\textbf{SwiGLU MLP}\\{\scriptsize $d_{\mathrm{ff}}{=}2304$}};
\node[plus, below=of mlp] (add3) {+};

\coordinate (in0) at ($(n1.north)+(0,6mm)$);
\coordinate (out0) at ($(add3.south)+(0,-6mm)$);

\draw[arrow] (in0) -- (n1.north);
\draw[arrow] (n1) -- (sa);
\draw[arrow] (sa) -- (add1);
\draw[arrow] (add1) -- (n2);
\draw[arrow] (n2) -- (mem);
\draw[arrow] (mem) -- (add2);
\draw[arrow] (add2) -- (n3);
\draw[arrow] (n3) -- (mlp);
\draw[arrow] (mlp) -- (add3);
\draw[arrow] (add3) -- (out0);

\node[cont, fit=(n1) (sa) (add1) (n2) (mem) (add2) (n3) (mlp) (add3),
      inner ysep=9pt] (blockbox) {};

\coordinate (rleft) at ($(blockbox.west)+(-14mm,0)$);



\coordinate (b0) at ($(n1.north)+(0,2.5mm)$);
\coordinate (b1) at ($($(add1.south)!0.45!(n2.north)$)+(0,0.8mm)$);
\coordinate (b2) at ($($(add2.south)!0.45!(n3.north)$)+(0,0.8mm)$);

\draw[resid] (b0) -- ($(rleft |- b0)$) |- (add1.west);
\draw[resid] (b1) -- ($(rleft |- b1)$) |- (add2.west);
\draw[resid] (b2) -- ($(rleft |- b2)$) |- (add3.west);

\node[note, rotate=90, color=orange!80!black] at ($(rleft)+(-2mm,0)$) {Residual};

\node[blk, fill=cRouteBox, right=24mm of n1] (rin) {\textbf{Router input}\\Pool($H$)};
\node[routedia, below=of rin] (router) {\textbf{Router}\\softmax($W_r r+b_r$)};
\node[blk, fill=cRouteBox, below=of router] (topk) {\textbf{Select Top-$k$}\\$k=64$};

\node[cont, fill=cBank, below=5mm of topk, minimum width=66mm, minimum height=28mm] (bank) {};
\node[bankhdr, anchor=north] at ($(bank.north)+(0mm,-1mm)$)
  {\textbf{Memory bank} \(M\)\\
   \(N_m=262{,}208,\ C=4{,}097,\ T=64\)};

\node[chap, fill=cRouteBox, anchor=west] (sh1) at ($(bank.west)+(4mm,2mm)$) {Sh1};
\node[chap, fill=blue!15, right=1.6mm of sh1] (r1) {R1};
\node[chap, fill=blue!15, right=1.1mm of r1] (r2) {R2};
\node[chap, draw=none, right=0.8mm of r2] (dots) {$\cdots$};
\node[chap, fill=blue!15, right=0.8mm of dots] (rk) {R64};
\node[chap, fill=gray!15, minimum width=52mm, below=6mm of sh1, anchor=west] (inactive) {4{,}032 inactive chapters};

\begin{scope}[on background layer]
  \node[cont, fit=(rin) (bank), fill=cRouteBG] (routebox) {};
\end{scope}
\node[note, anchor=south] at ($(routebox.north)+(0mm,1mm)$) {\textbf{Chapter-based memory routing}};

\draw[arrow] (rin) -- (router);
\draw[arrow] (router) -- (topk);
\draw[arrow] (topk) -- (bank);

\draw[routearrow]
  (n2.east) to[out=8, in=188, looseness=1.15]
  node[pos=0.30, above, yshift=5pt, xshift=-2pt, lab, text=orange!80!black] {Router Input}
  ([xshift=-1.5mm]rin.west);

\draw[kv]
  ([yshift=-2mm]bank.west) to[out=195, in=15, looseness=1.08]
  node[pos=0.33, below, yshift=-6pt, xshift=1pt, lab, text=red!70!black] {$K,V$}
  ([xshift=1mm]mem.east);

\end{tikzpicture}
\caption{\textbf{Single transformer block with chapter-based memory routing}}
\label{fig:memory_arch_tikz}
\end{figure*}
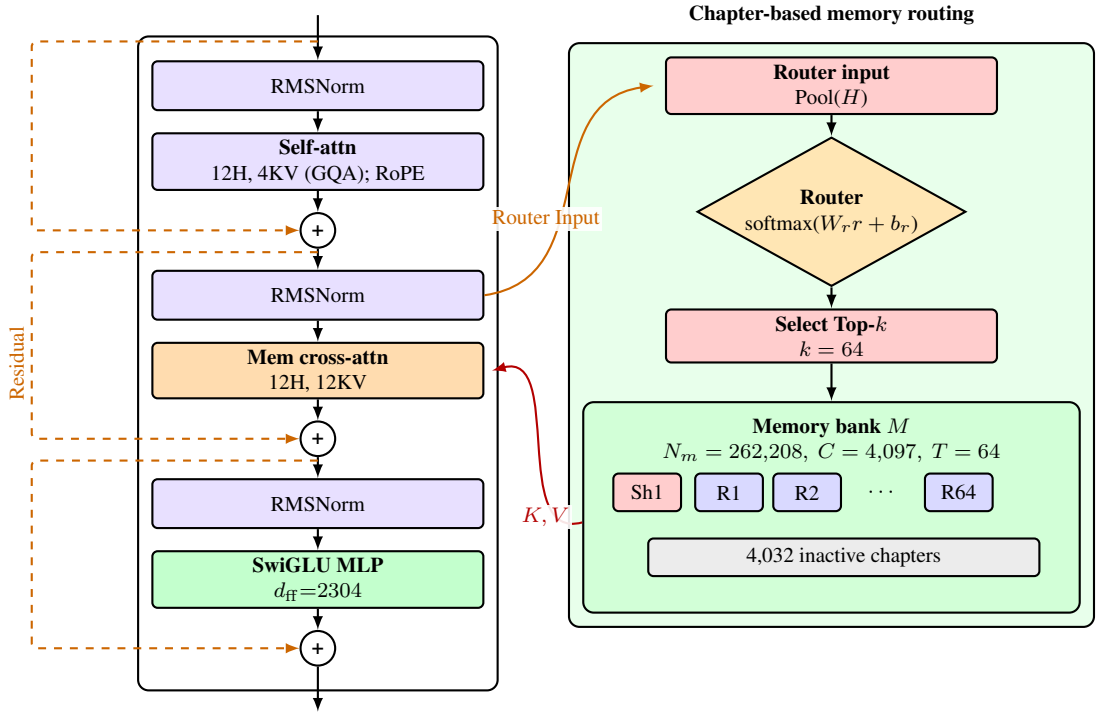

\section{Method}
\label{sec:method}

\subsection{Memory layer: learned bank with chapter routing}
We augment a standard decoder-only transformer \citep{vaswani2017attention} with a \emph{memory layer} that provides persistent, addressable storage. The memory consists of a learned bank of latent tokens that are trained end-to-end, and accessed through cross-attention, following the general memory-augmented attention pattern \citep{sukhbaatar2019memory,wu2020memformer}.

Let $H \in \mathbb{R}^{B \times L \times d}$ be the token hidden states at a layer. We maintain a memory bank
\[
M \in \mathbb{R}^{N_m \times d},
\]
where each row is a learned memory token. The memory layer reads from $M$ using cross-attention where the token stream produces queries and memory produces keys and values:
\begin{align}
Q &= \mathrm{Norm}(H) W_Q, \\
K &= M W_K, \\
V &= M W_V, \\
\mathrm{MemRead}(H,M) &= \mathrm{softmax}\!\left(\frac{QK^\top}{\sqrt{d_h}}\right)V.
\end{align}
The readout is added back to the token stream through a residual connection:
\[
H' = H + \, \mathrm{MemRead}(H,M)
\]

\subsection{Scaling the bank with chapters}
Attending over all $N_m$ memory tokens can be expensive for large banks. We therefore partition the bank into $C$ \emph{chapters}:
\[
M = [M_1; \ldots; M_C], \quad M_c \in \mathbb{R}^{T \times d}, \quad N_m = C T.
\]
For each input sequence, a lightweight router selects a small subset of chapters. We compute a sequence representation by pooling hidden states (for example, mean pooling):
\[
r = \mathrm{Pool}(H) \in \mathbb{R}^{d},
\]
then score chapters and select the top-$k$:
\[
p = \mathrm{softmax}(W_r r + b_r), \quad \mathcal{S} = \mathrm{TopK}(p,k).
\]
The memory layer then attends only to the selected chapters:
\[
M_{\mathcal{S}} = \mathrm{Concat}(\{M_c: c \in \mathcal{S}\}), \qquad
H' = H +  \mathrm{MemRead}(H, M_{\mathcal{S}}).
\]
This reduces memory attention cost from $O(LN_m)$ to $O(LkT)$ per memory layer while preserving attention-based associative retrieval. Algorithm~\ref{alg:moc_forward} (see Appendix~\ref{sec:alg_moc}) summarizes the forward pass of a single MoC memory layer, including sequence-level routing, chapter selection, and routed memory cross-attention. The routing mechanism is inspired by sparse conditional computation in mixture-of-experts models \citep{shazeer2017moe,fedus2022switch} and routing-based sparsification \citep{roy2021routing}.

\section{Experiments}
\label{sec:experiments}

\subsection{Setup}
\paragraph{Models.}
We compare (i) \textbf{Vanilla (iso-FLOP)}: a dense transformer baseline compute-matched to our memory model during pretraining (see Appendix~\ref{sec:analytic_flops} for the analytic FLOPs calculation), (ii) \textbf{Mixture of Chapters (MoC)}: our memory-augmented transformer, and (iii) \textbf{Vanilla (backbone-only)}: the dense transformer backbone of the memory model with memory components removed.

\paragraph{Training protocol.}
We pretrain for 9{,}600 steps (9.6B tokens) and then instruction fine-tune (IFT) for 2 epochs on 230M tokens (3.2k steps). This is a relatively heavy post-training budget for this model scale and it induces clear forgetting in the dense baseline. During IFT, we increase context length from 1024 to 2048. All plots and reported pretraining results correspond to the 9{,}600-step run (not a continued run).

\paragraph{Evaluation.}
We report accuracy on MMLU \citep{hendrycks2021mmlu}, ARC-Challenge \citep{clark2018arc}, BoolQ \citep{clark2019boolq}, and OpenBookQA \citep{mihaylov2018openbookqa}. Random guessing is approximately 0.25 on ARC-Challenge and 0.50 on BoolQ.

\subsection{Results}
\begin{figure*}[t]
\centering
\begin{minipage}{0.49\textwidth}
  \centering
  \includegraphics[width=\linewidth]{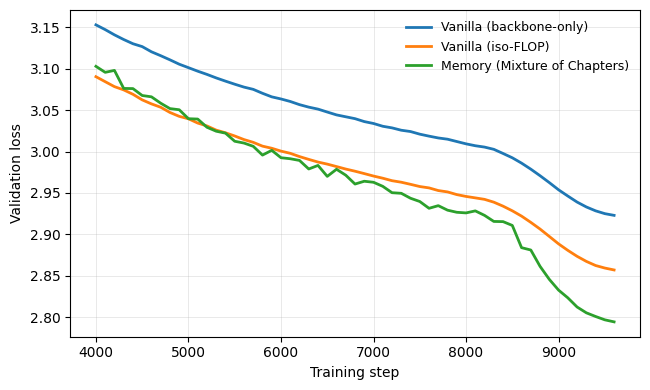}

  {\small (Left) Pretraining eval loss (9{,}600 steps).}
\end{minipage}\hfill
\begin{minipage}{0.49\textwidth}
  \centering
  \includegraphics[width=\linewidth]{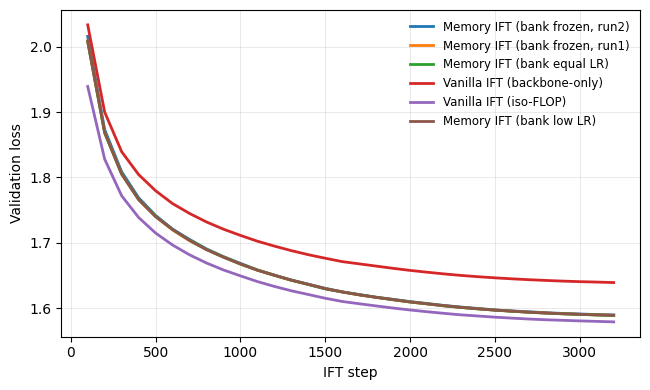}

  {\small (Right) IFT eval loss for bank-freeze and LR variants.}
\end{minipage}
\caption{Loss curves. Left: Pretraining. Right: IFT (Freezing the bank yields identical curves).}
\label{fig:loss-curves}
\end{figure*}

\paragraph{Pretraining loss.}
Table~\ref{tab:pretrain-loss} shows validation loss at the end of pretraining. The memory model attains the best loss, and Figure~\ref{fig:loss-curves} shows it continues to separate late in training, suggesting additional headroom with longer runs.

\begin{table}[t]
\centering
\caption{Pretraining validation loss (lower is better).}
\label{tab:pretrain-loss}
\begin{tabular}{l c}
\toprule
Model & Val loss $\downarrow$ \\
\midrule
Vanilla (backbone-only) & 2.92 \\
Vanilla (iso-FLOP) & 2.86 \\
Mixture of Chapters (MoC) & \textbf{2.79} \\
\bottomrule
\end{tabular}
\end{table}

\paragraph{Benchmarks and forgetting under heavy IFT.}
Table~\ref{tab:bench-all} reports benchmark accuracy after pretraining and after IFT, along with deltas (IFT minus pretrain).
Vanilla is strongly affected on knowledge-heavy tasks: ARC-Challenge drops by 0.0669 (0.3177 $\rightarrow$ 0.2508, approaching random) and BoolQ drops by 0.0624 (0.5713 $\rightarrow$ 0.5089, approaching random).
In contrast, the memory model remains stable on these tasks: BoolQ is effectively unchanged (+0.0024), and ARC-Challenge drops only 0.0268, less than half the vanilla degradation. This pattern is consistent with the memory bank anchoring factual knowledge during continued training.


\begin{table*}[t]
\centering
\small
\setlength{\tabcolsep}{7pt}
\caption{Comparison of Vanilla, Mixture of Chapters (MoC) ($\Delta$ = IFT - pretrain; lower is better)}
\label{tab:bench-all}
\begin{tabular}{l ccc ccc}
\toprule
& \multicolumn{3}{c}{Vanilla (iso-FLOP)} & \multicolumn{3}{c}{MoC} \\
\cmidrule(lr){2-4} \cmidrule(lr){5-7}
Benchmark & Pretrain (\%) & $\Delta$ (pp) $\downarrow$ & IFT (\%) & Pretrain (\%) & $\Delta$ (pp) $\downarrow$ & IFT (\%) \\
\midrule
MMLU  & 26.90 & -0.99 & 25.91 & \textbf{27.87} & \textbf{-0.35} & \textbf{27.52} \\
ARC-C & \textbf{31.77} & -6.69 & 25.08 & 31.44 & \textbf{-2.68} & \textbf{28.76} \\
BoolQ & 57.13 & -6.24 & 50.89 & \textbf{61.87} & \textbf{+0.24} & \textbf{62.11} \\
OBQA  & 36.20 & -2.00 & 34.20 & \textbf{37.40} & -2.00 & \textbf{35.40} \\
\bottomrule
\end{tabular}
\end{table*}

\paragraph{IFT loss and freezing the memory bank.}
Figure~\ref{fig:loss-curves} (right) compares IFT validation loss for memory variants where the bank is frozen, trained with a very small learning rate, or trained with the same learning rate as the backbone. The curves overlap closely, and post-IFT benchmark scores remain within noise across these settings (Appendix~\ref{sec:exp_appendix}). This indicates the pretrained memory bank can be reused during post-training without further bank updates.

\paragraph{Pretraining distribution after IFT.}
Evaluating IFT checkpoints on the pretraining validation distribution yields higher loss for the memory model (3.2024) than vanilla (3.0574), despite better retention on knowledge benchmarks. This suggests the memory model can shift further toward instruction alignment while preserving factual knowledge, consistent with a separation of roles between backbone adaptation and memory anchoring.

\section{Conclusion}
We introduced \textbf{Mixture of Chapters (MoC)}, a memory-augmented transformer that adds a learned latent memory bank and scales access via sequence-level chapter routing, providing an explicit associative memory substrate with sparse, tractable access.\\

In a 9.6B-token pretraining run, the memory model achieves lower validation loss than compute-matched dense baselines and improves knowledge-heavy benchmark performance. Under heavy post-training (2-epoch IFT on 230M tokens with context length increased from 1024 to 2048), the dense baseline exhibits substantial forgetting on ARC-Challenge and BoolQ, while the memory model remains comparatively stable; freezing the bank during IFT performs on par with updating it. Overall, these results point to learned memory as a complementary axis of scaling and motivate further study of larger banks, longer training, and improved routing/organization for specialization and interpretability.

\newpage
\bibliography{main}
\bibliographystyle{plainnat}

\newpage
\appendix
\section{Appendix}
\label{sec:appendix}

\subsection{Extended literature review (concise map)}
\label{sec:litreview}
Due to the page limit, the main paper focuses on the closest comparisons (Titans, Memory Layers at Scale, PKM). This appendix provides a compact narrative overview of broader related directions, complemented by Table~\ref{tab:lit_map}.

\paragraph{Inference-time and external memory.}
A common strategy is to externalize memory at inference time, either by caching past activations to improve efficiency and extend effective context, or by retrieving text from an external index. Transformer-XL is a canonical example of recurrence and caching \citep{dai2019txl}, while RAG exemplifies retrieval-conditioned generation \citep{lewis2020rag}. These approaches are strong baselines, but the stored content remains outside the model and must be fetched and integrated per query.

\paragraph{Learned memory inside the model.}
A complementary line integrates memory as a trainable component. Memformer reads and writes through an internal memory mechanism coupled to token representations \citep{wu2020memformer}. For scaling, PKM provides efficient large key-value lookup \citep{lample2019pkm}, and Memory Layers at Scale shows that trainable memory layers can add substantial capacity with near-constant FLOPs and strong factual gains \citep{berges2025memorylayers}. Our work is closest to this direction, but uses a learned latent-token bank accessed by cross-attention and scales access via sequence-level chapter routing.

\paragraph{Updatable and test-time memory.}
Some methods emphasize explicit test-time updates or state that evolves across interactions, such as Titans \citep{behrouz2025titans}. These approaches often target persistent personalization or long-horizon interaction, whereas our focus is on scaling learned associative memory trained end-to-end and analyzing retention under continued training.

\paragraph{Sparse routing and conditional computation.}
Routing and conditional computation provide a general template for scaling by activating only a subset of components. MoE systems route inputs to a small subset of experts \citep{shazeer2017moe,fedus2022switch}. Our method borrows this principle but applies routing to memory selection, with chapters serving as routed memory units.

\paragraph{Retention and catastrophic forgetting.}
Catastrophic forgetting is a classical issue in sequential training \citep{kirkpatrick2017ewc}, and has been measured in continual or sequential tuning of language models. Our experiments connect this literature to architectural memory by testing whether a learned bank helps preserve factual knowledge across the pretraining-to-IFT transition.

\begin{table}[!htbp]
\caption{Literature map (representative, not exhaustive).}
\label{tab:lit_map}
\centering
\scriptsize
\setlength{\tabcolsep}{4pt}
\renewcommand{\arraystretch}{1.1}
\begin{tabular}{p{0.30\linewidth}p{0.66\linewidth}}
\toprule
Category & Representative works \\
\midrule
Learnable internal memory bank (cross-attn / persistent tokens) &
Memformer~\citep{wu2020memformer}, LM2~\citep{kang2025lm2}, persistent memory vectors~\citep{sukhbaatar2019persistent}, stateful memory variants~\citep{wu2022stateful}. \\

Large trainable memory layers (scalable access) &
Memory Layers at Scale~\citep{berges2025memorylayers}, PKM~\citep{lample2019pkm}, EMAT~\citep{wu2022emat}, Memorizing Transformers~\citep{wu2022memorizing}. \\

Test-time / updatable memory &
Titans~\citep{behrouz2025titans}, MemoryLLM~\citep{wang2024memoryllm}, M+~\citep{wang2025mplus}, Larimar~\citep{liu2024larimar}, EM-LLM~\citep{fountas2025emllm}, Engram~\citep{deepseek2026engram}. \\

Recurrence and compressive memory for long context &
Transformer-XL~\citep{dai2019txl}, RMT~\citep{bulatov2022rmt,bulatov2023scaling}, Compressive Transformers~\citep{rae2020compressive}, Infini-attention~\citep{munkhdalai2024infini}, Cached Transformers~\citep{zhang2023cached}. \\

Structured/entity memories &
Mention Memory~\citep{dearmas2021mention}, Entities as Experts~\citep{fevry2020entities}, Facts as Experts~\citep{verga2020facts}. \\

Latent-token architectures and context compression &
Perceiver/Perceiver IO~\citep{jaegle2021perceiver,jaegle2021perceiverio}, Flamingo~\citep{alayrac2022flamingo}, Activation Beacon~\citep{activationbeacon2024}, ICAE~\citep{icae2023}, compressed-context memory~\citep{kim2024ccm}. \\

Retrieval and KV-cache baselines &
RAG~\citep{lewis2020rag}, RETRO~\citep{borgeaud2021retro}, kNN-LM~\citep{khandelwal2019knnlm}, KV-cache retention/compression (H$_2$O, StreamingLLM, NACL, Trellis)~\citep{zhang2023h2o,xiao2023streamingllm,chen2024nacl,karami2025trellis}. \\
\bottomrule
\end{tabular}
\end{table}

\subsection{Analytic FLOPs calculation}
\label{sec:analytic_flops}

In this section, we estimate FLOPs analytically for a \emph{single sequence} with batch size \(B=1\) and sequence length \(L=1024\). Our codebase includes a reference implementation, \texttt{estimate\_flops.py}, which follows the same counting rules used here for complete training runs. The derivation below mirrors that implementation, while making the layer-by-layer calculation explicit. 

Linear layers use
\[
\mathrm{FLOPs}_{\mathrm{linear}} = 2 \, N \, d_{\mathrm{in}} \, d_{\mathrm{out}},
\]
attention matmuls use
\[
\mathrm{FLOPs}_{\mathrm{attn\ matmul}} = 4 \, B \, L_q \, L_k \, d,
\]
RMSNorm uses
\[
\mathrm{FLOPs}_{\mathrm{RMSNorm}} = N (4d + 4),
\]
and the backward pass is approximated as \(2\times\) the forward pass, consistent with the estimator.

For our backbone, \(d=768\), \(d_{\mathrm{ff}}=2304\), \(h=12\), \(h_{\mathrm{kv}}=4\), so the self-attention KV dimension is
\[
d_{\mathrm{kv}} = h_{\mathrm{kv}} \cdot \frac{d}{h} = 4 \cdot 64 = 256.
\]

\paragraph{Standard transformer layer.}
A standard layer contains self-attention, two RMSNorms, one SwiGLU MLP, RoPE, and two residual additions.

\textbf{Self-attention projections and attention.}
\begin{align}
Q &: 2 \cdot 1024 \cdot 768 \cdot 768 = 1{,}207{,}959{,}552 \\
K &: 2 \cdot 1024 \cdot 768 \cdot 256 = 402{,}653{,}184 \\
V &: 2 \cdot 1024 \cdot 768 \cdot 256 = 402{,}653{,}184 \\
O &: 2 \cdot 1024 \cdot 768 \cdot 768 = 1{,}207{,}959{,}552 \\
QK^\top,\ AV &: 4 \cdot 1 \cdot 1024 \cdot 1024 \cdot 768 = 3{,}221{,}225{,}472.
\end{align}
The estimator also includes attention softmax, masking, and scaling:
\[
12 \cdot 1024 \cdot 1024 \cdot (5+2) = 88{,}080{,}384.
\]
Hence total self-attention FLOPs per standard layer are
\[
6{,}530{,}531{,}328.
\]

\textbf{RoPE.}
\[
3 \cdot 1024 \cdot (768 + 256) = 3{,}145{,}728.
\]

\textbf{Two RMSNorms.}
\[
2 \cdot 1024 \cdot (4\cdot 768 + 4) = 6{,}299{,}648.
\]

\textbf{SwiGLU MLP.}
\begin{align}
W_{\mathrm{up}} &: 2 \cdot 1024 \cdot 768 \cdot 2304 = 3{,}623{,}878{,}656 \\
W_{\mathrm{gate}} &: 2 \cdot 1024 \cdot 768 \cdot 2304 = 3{,}623{,}878{,}656 \\
W_{\mathrm{down}} &: 2 \cdot 1024 \cdot 2304 \cdot 768 = 3{,}623{,}878{,}656 \\
\text{SwiGLU activation} &: 1024 \cdot 2304 \cdot 5 = 11{,}796{,}480.
\end{align}
So total MLP FLOPs per layer are
\[
10{,}883{,}432{,}448.
\]

\textbf{Residual adds.}
\[
2 \cdot 1024 \cdot 768 = 1{,}572{,}864.
\]

Putting these together, one \emph{standard} transformer layer costs
\[
6{,}530{,}531{,}328
+ 3{,}145{,}728
+ 6{,}299{,}648
+ 10{,}883{,}432{,}448
+ 1{,}572{,}864
= 17{,}424{,}982{,}016
\]
FLOPs.

\paragraph{MoC memory layer.}
An MoC memory layer adds router computation, memory preprocessing, memory cross-attention, an extra RMSNorm, and an extra residual add.

The memory bank has \(262{,}208\) tokens, partitioned into \(4097\) chapters, giving
\[
T = 262{,}208 / 4097 = 64
\]
tokens per chapter. With one shared chapter and top-\(k=64\) routed chapters, the selected memory tokens per sequence are
\[
N_{\mathrm{sel}} = (1 + 64)\cdot 64 = 4160.
\]

\textbf{Router forward.}
For sequence-level routing, the estimator includes sequence pooling, router projection, softmax, and top-\(k\):
\begin{align}
\text{pooling} &: 1 \cdot 768 \cdot (1024-1) + 768 = 786{,}432 \\
\text{router linear} &: 2 \cdot 1 \cdot 768 \cdot 4097 = 6{,}292{,}992 \\
\text{router softmax} &: 4097 \cdot 5 = 20{,}485 \\
\text{top-}k &: 4097 \cdot \lceil \log_2 64 \rceil = 24{,}582.
\end{align}
Thus router forward FLOPs are
\[
7{,}124{,}491.
\]

\textbf{Router auxiliary-loss compute.}
The estimator also includes the auxiliary routing-loss terms (load-balancing, z-loss, entropy, and related reductions), which contribute
\[
331{,}859
\]
FLOPs per memory layer.

\textbf{Memory preprocessing.}
The estimator includes chapter weighting plus normalization of the selected memory tokens:
\begin{align}
\text{weighting} &: 1 \cdot 4160 \cdot 768 = 3{,}194{,}880 \\
\text{RMSNorm} &: 4160 \cdot (4\cdot 768 + 4) = 12{,}796{,}160.
\end{align}
So memory preprocessing contributes
\[
15{,}991{,}040
\]
FLOPs.

\textbf{Memory attention.}
Here \(h_{\mathrm{mem}} = 12\), \(h_{\mathrm{mem,kv}}=12\), so memory KV dimension is \(768\).
\begin{align}
Q &: 2 \cdot 1024 \cdot 768 \cdot 768 = 1{,}207{,}959{,}552 \\
K &: 2 \cdot 4160 \cdot 768 \cdot 768 = 4{,}907{,}335{,}680 \\
V &: 2 \cdot 4160 \cdot 768 \cdot 768 = 4{,}907{,}335{,}680 \\
O &: 2 \cdot 1024 \cdot 768 \cdot 768 = 1{,}207{,}959{,}552 \\
QK^\top,\ AV &: 4 \cdot 1 \cdot 1024 \cdot 4160 \cdot 768 = 13{,}086{,}351{,}360.
\end{align}
The attention softmax, masking, and scaling term is
\[
12 \cdot 1024 \cdot 4160 \cdot (5+2) = 357{,}826{,}560.
\]
Hence memory attention contributes
\[
25{,}674{,}645{,}504
\]
FLOPs.

\textbf{Extra norm and residual.}
\[
\text{extra RMSNorm} = 1024 \cdot (4\cdot 768 + 4) = 3{,}149{,}824,
\qquad
\text{extra residual} = 1024 \cdot 768 = 786{,}432.
\]

Therefore the \emph{extra} cost added by an MoC memory layer is
\[
7{,}124{,}491
+ 331{,}859
+ 15{,}991{,}040
+ 25{,}674{,}645{,}504
+ 3{,}149{,}824
+ 786{,}432
= 25{,}702{,}029{,}150.
\]
Adding the underlying standard transformer layer gives
\[
17{,}424{,}982{,}016 + 25{,}702{,}029{,}150
= 43{,}127{,}011{,}166
\]
FLOPs per MoC memory layer.

\paragraph{Final prediction head and loss.}
After the transformer stack, the estimator includes a final RMSNorm, LM head, and cross-entropy loss:
\begin{align}
\text{final RMSNorm} &: 1024 \cdot (4\cdot 768 + 4) = 3{,}149{,}824 \\
\text{LM head} &: 2 \cdot 1024 \cdot 768 \cdot 49152 = 77{,}309{,}411{,}328 \\
\text{CE loss} &: (1024-1)\cdot 49152 \cdot 5 = 251{,}412{,}480.
\end{align}
Thus the final prediction/loss block contributes
\[
77{,}563{,}973{,}632
\]
FLOPs.

\paragraph{Model totals.}
The backbone-only vanilla model has 16 standard layers:
\[
F_{\mathrm{fwd}}^{\mathrm{backbone}}
= 16 \cdot 17{,}424{,}982{,}016 + 77{,}563{,}973{,}632
= 356{,}363{,}685{,}888.
\]
The iso-FLOP vanilla baseline has 24 standard layers:
\[
F_{\mathrm{fwd}}^{\mathrm{iso}}
= 24 \cdot 17{,}424{,}982{,}016 + 77{,}563{,}973{,}632
= 495{,}763{,}542{,}016.
\]
The memory model has 12 standard layers and 4 MoC memory layers:
\[
F_{\mathrm{fwd}}^{\mathrm{MoC}}
= 12 \cdot 17{,}424{,}982{,}016
+ 4 \cdot 43{,}127{,}011{,}166
+ 77{,}563{,}973{,}632
= 459{,}171{,}802{,}488.
\]

Following the estimator, we approximate one backward pass as \(2\times\) the forward FLOPs:
\[
F_{\mathrm{bwd}} \approx 2F_{\mathrm{fwd}}.
\]
Hence the total for one forward plus one backward pass is \(3F_{\mathrm{fwd}}\).

\begin{table}[t]
\centering
\small
\caption{Per-sequence FLOPs for one forward pass and one forward+backward pass (\(B=1, L=1024\)).}
\label{tab:analytic_flops}
\begin{tabular}{lccc}
\toprule
Model & Forward FLOPs & Backward FLOPs & Forward+Backward FLOPs \\
\midrule
Vanilla (backbone-only) & 0.356T & 0.713T & 1.069T \\
Vanilla (iso-FLOP) & 0.496T & 0.992T & 1.487T \\
Mixture of Chapters (MoC) & 0.459T & 0.918T & 1.378T \\
\bottomrule
\end{tabular}
\end{table}

Note that these are \emph{single-sequence} estimates. They differ slightly from dividing the microstep estimates by batch size, because the sequence-level router auxiliary-loss terms do not scale exactly linearly with batch size.

\subsection{Algorithm: Mixture of Chapters (MoC) memory layer}
\label{sec:alg_moc}

\begin{algorithm}[h]
\caption{MoC memory layer with sequence-level routing}
\label{alg:moc_forward}
\begin{algorithmic}[1]
\Require Batch hidden states \(H \in \mathbb{R}^{B \times L \times d}\)
\Require Chaptered memory bank \(M = [M_1; \dots; M_C]\), where \(M_c \in \mathbb{R}^{T \times d}\)
\Require Router parameters \(W_r, b_r\), projections \(W_Q, W_K, W_V\), top-\(k\)
\Ensure Updated hidden states \(H' \in \mathbb{R}^{B \times L \times d}\)

\For{\(b = 1\) to \(B\)}
    \State \(r_b \gets \mathrm{Pool}(H_b)\)
    \State \(p_b \gets \mathrm{softmax}(W_r r_b + b_r)\)
    \State \(\mathcal{S}_b \gets \mathrm{TopK}(p_b, k)\)
    \State \(M_b \gets \mathrm{Concat}(\{M_c : c \in \mathcal{S}_b\})\)

    \State \(Q_b \gets \mathrm{Norm}(H_b) W_Q\)
    \State \(K_b \gets M_b W_K\)
    \State \(V_b \gets M_b W_V\)

    \State \(A_b \gets \mathrm{softmax}\!\left(\frac{Q_b K_b^\top}{\sqrt{d_h}}\right)\)
    \State \(\mathrm{MemRead}_b \gets A_b V_b\)
    \State \(H'_b \gets H_b + \mathrm{MemRead}_b\)
\EndFor

\State \Return \(H' = [H'_1; \dots; H'_B]\)
\end{algorithmic}
\end{algorithm}

\subsection{Additional experimental details}
\label{sec:exp_appendix}
\subsubsection{Model sizes}
The memory model has a larger total parameter count due to the persistent bank, but uses sparse chapter routing so only a subset of the bank is activated per input. We therefore match pretraining compute to a dense baseline (Vanilla iso-FLOP). For the non-memory baselines, the \textbf{Vanilla (backbone-only)} model uses the same 16-layer backbone architecture with the memory layers removed, while the \textbf{Vanilla (iso-FLOP)} baseline is a denser variant obtained by repeating the standard backbone layers without memory modules and increasing depth to match the pretraining compute budget of the memory model. For reference, the dense baseline has 202.94M parameters, while the memory model has 371.29M total parameters composed of a 147.87M backbone plus a 201.38M bank and 22.04M memory-layer parameters.

\subsubsection{Model and training configuration}
\label{app:config}

Tables~\ref{tab:arch-config} and \ref{tab:train-config} summarize the main Mixture of Chapters (MoC) architecture and the training settings used for pretraining and instruction fine-tuning (IFT).

\begin{table*}[t]
\centering
\small
\caption{Main MoC architecture used in our experiments.}
\label{tab:arch-config}
\begin{tabular}{p{0.28\linewidth} p{0.64\linewidth}}
\toprule
Component & Setting \\
\midrule
Backbone &
Decoder-only transformer with 16 layers, hidden size 768, 12 attention heads, 4 KV heads, SwiGLU MLP with intermediate size 2304, RMSNorm, RoPE with $\theta=100000$, tied embeddings, vocabulary size 49152 \\
Memory placement &
Memory cross-attention inserted at layers $\{2,6,10,14\}$ \\
Memory bank &
262{,}208 latent memory tokens, shared across memory layers \\
Chapter routing &
4097 total chapters, including 1 shared chapter; top-$k=64$ chapters selected per sequence \\
Memory attention &
12 memory query heads and 12 memory KV heads \\
Routing details &
Sequence-level routing, routed scaling factor 2.5, shared/routed mixing normalized with RMSNorm \\
Regularization &
Load-balance loss coefficient 0.01, $z$-loss coefficient 0.001 \\
Other &
Memory adapter enabled, memory initialization std.\ 0.02, memory quantization disabled \\
\bottomrule
\end{tabular}
\end{table*}

\begin{table*}[t]
\centering
\small
\setlength{\tabcolsep}{5pt}
\renewcommand{\arraystretch}{1.12}
\caption{Main training settings for pretraining and IFT.}
\label{tab:train-config}
\begin{tabularx}{\textwidth}{>{\raggedright\arraybackslash}p{0.18\textwidth} Y Y}
\toprule
Setting & Pretraining & IFT \\
\midrule
Tokenizer & \path{HuggingFaceTB/SmolLM2-135M} & \path{HuggingFaceTB/SmolLM2-135M-Instruct} \\
Dataset & \path{Tasmay-Tib/fineweb-edu-10bt-split} & \path{HuggingFaceH4/ultrachat_200k} \\
Data split & train / eval & train\_sft / test\_sft \\
Sequence length & 1024 & 2048 \\
Distributed setup & DDP on 8 GPUs & DDP on 4 GPUs \\
Batch size & 32 with gradient accumulation 4 & 32 with gradient accumulation 1 \\
Training duration & 9600 steps & 2 epochs (3200 steps) \\
Optimizer & AdamW & AdamW \\
Learning rates &
base model: $3 \times 10^{-4}$ \newline
memory layers: $6 \times 10^{-4}$ \newline
memory bank: $6 \times 10^{-4}$ &
base model: $3 \times 10^{-5}$ \newline
memory layers: $1.5 \times 10^{-5}$ \newline
memory bank: $5 \times 10^{-6}$ \\
Scheduler &
WSD, warmup 250 steps, min LR ratio 0.1, decay start at step 8160 &
Cosine, warmup 250 steps \\
Optimization details &
weight decay 0.1, Adam $\beta_1=0.9$, $\beta_2=0.95$, grad clip 1.0 &
weight decay 0.1, Adam $\beta_1=0.9$, $\beta_2=0.95$, grad clip 1.0 \\
Precision & bf16 & bf16 \\
Checkpointing & gradient checkpointing enabled & gradient checkpointing enabled \\
Initialization & random initialization & initialized from the pretrained MoC checkpoint \\
\bottomrule
\end{tabularx}
\end{table*}

For comparison, the vanilla backbone-only baseline uses the same 16-layer backbone with memory disabled, while the iso-FLOP vanilla baseline increases depth to 24 layers.

\subsubsection{Pretraining benchmarks including the backbone-only baseline}
Table~\ref{tab:pretrain-bench-appendix} includes the backbone-only benchmark results to contextualize gains from MoC versus simply changing backbone capacity.

\begin{table}[t]
\centering
\caption{Benchmark accuracy after pretraining (including backbone-only).}
\label{tab:pretrain-bench-appendix}
\begin{tabular}{l c c c c}
\toprule
Model & MMLU & ARC-C & BoolQ & OBQA \\
\midrule
Vanilla (iso-FLOP) & 0.2690 & 0.3177 & 0.5713 & 0.3620 \\
MoC & \textbf{0.2787} & 0.3144 & \textbf{0.6187} & \textbf{0.3740} \\
Vanilla (backbone-only) & 0.2746 & \textbf{0.3311} & 0.5446 & 0.3340 \\
\bottomrule
\end{tabular}
\end{table}

\subsubsection{IFT learning rates and bank update ablations}
IFT uses a base learning rate of $3\times 10^{-5}$ (one tenth of the pretraining rate). We vary the memory bank learning rate across:
(1) \textbf{Frozen bank} (LR = 0),
(2) \textbf{Low bank LR} (very small LR),
(3) \textbf{Equal LR} (bank LR matches base LR).
Across these settings, Figure~\ref{fig:loss-curves} shows no visible divergence in validation loss curves.

Table~\ref{tab:ift-bank-ablations} reports benchmark scores for representative settings. Freezing the bank performs on par with updating it, indicating that the pretrained bank is retained and sufficient for post-training.

\begin{table}[t]
\centering
\caption{MoC IFT benchmarks for bank update settings.}
\label{tab:ift-bank-ablations}
\begin{tabular}{l c c c c}
\toprule
IFT setting & MMLU & ARC-C & BoolQ & OBQA \\
\midrule
MoC (bank frozen) & \textbf{0.2754} & 0.2843 & \textbf{0.6214} & \textbf{0.3540} \\
\textbf{MoC (bank 1/10 LR)$^*$} & 0.2752 & 0.2876 & 0.6211 & \textbf{0.3540} \\
MoC (bank equal LR) & 0.2753 & \textbf{0.2943} & 0.6211 & 0.3480 \\
\bottomrule
\end{tabular}

\vspace{0.25em}
{\centering \textit{$^*$ denotes the main setting reported above.}\par}
\end{table}

\subsubsection{IFT shifts the pretraining distribution loss while preserving knowledge benchmarks}
When evaluating IFT checkpoints on the pretraining validation distribution, the memory model exhibits higher loss (3.2024) than the vanilla baseline (3.0574), even though it retains benchmark performance substantially better on ARC-Challenge and BoolQ. This suggests that the memory model can adapt more strongly toward the IFT objective while keeping factual knowledge stable through the explicit bank, consistent with a division of labor between backbone adaptation and memory anchoring.

\subsubsection{Compute and memory implications during post-training}
Freezing the memory bank during IFT removes bank updates and reduces optimizer state and gradient memory for that component, which lowers VRAM requirements. In addition, increasing context length from 1024 to 2048 increases self-attention cost for all models. Under this longer-context setting, the benefit of freezing the bank is preserved while the dense baseline pays the full quadratic self-attention cost. This combination makes the memory approach attractive for heavier or longer post-training schedules.

\subsection{Ablation studies and future work}
\label{sec:ablations_future}

This section lists ablations for \emph{Mixture of Chapters} and outlines directions to clarify scaling, efficiency, and retention.

\paragraph{Ablation studies.}
\begin{itemize}
    \item \textbf{Top-$k$ routing sweep.}
    Vary the number of routed chapters, e.g., $k \in \{1,2,4,8,16,32,64\}$, holding bank size fixed to map the quality--compute frontier.

    \item \textbf{Routing granularity and train/inference mismatch.}
    Compare (i) sequence routing at train and inference (current), (ii) token routing at inference only, and (iii) token routing at train and inference (if feasible).

    \item \textbf{Shared chapters.}
    Sweep the number of always-on shared chapters and mixing schemes:
    $n_{\text{shared}}\in\{0,1,2,4,8\}$.

    \item \textbf{Memory placement and sharing.}
    Study where memory layers are inserted and whether the bank is shared across layers vs per-layer banks.

    \item \textbf{Router regularization.}
    Sweep load-balancing / auxiliary routing losses (and any $z$-loss) to quantify utilization collapse vs quality tradeoffs.

    \item \textbf{Initialization.}
    Ablate output-projection initialization (e.g., zero-init) and bank initialization scale to test stability and early training behavior.

    \item \textbf{Memory size and chapter geometry.}
    Scale total memory tokens, number of chapters, and tokens-per-chapter $T$ to separate “more chapters” from “longer chapters.”

    \item \textbf{Compression ablations.}
    Evaluate quantized banks and low-rank factorization of bank/projections to trade expressiveness for capacity.

    \item \textbf{PEFT comparisons: LoRA vs memory adapters vs both.}
    Compare LoRA-only, memory-adapter-only, and combined settings under matched trainable parameter budgets.
\end{itemize}

\paragraph{Future work.}
\begin{itemize}
    \item \textbf{Scaling laws for learned memory.}
    Extend pretraining beyond 9.6B tokens and sweep bank capacity to quantify how performance scales with memory under fixed compute.

    \item \textbf{Token-level routing with efficient kernels.}
    Investigate kernels that avoid materializing large routed key tensors, enabling token routing during training without large memory overhead.

    \item \textbf{Dynamic and test-time updates.}
    Add controlled write/update mechanisms to support continual acquisition and personalization while retaining the “freeze bank” benefits.

    \item \textbf{Interpretability and chapter-level editing.}
    Analyze router selections for specialization, and test targeted chapter edits (zeroing, swapping, pruning) to localize knowledge.

    \item \textbf{Hybrid memory: shared global + routed private.}
    Explore explicit separation between a small shared global memory and a large routed bank, with different update schedules or learning rates.

    \item \textbf{Broader evaluations.}
    Add tests for long-context generalization, factual recall, and retention under multiple sequential post-training stages.

    \item \textbf{Interpretability and analysis of memory usage.}
    Analyze how memory is used across layers and time: (i) chapter selection statistics (entropy, sparsity, stability across prompts), (ii) per-layer reliance on memory (attention mass to memory vs self-attn), (iii) chapter specialization (clustering chapters by the queries they serve or by input domains), and (iv) causal interventions such as zeroing/swapping/pruning selected chapters or memory tokens to localize which knowledge is stored where. This can be paired with qualitative retrieval probes by decoding nearest-neighbor token projections from memory to inspect what each chapter appears to represent.

\end{itemize}

\end{document}

%% file: math_commands.tex
\usepackage{amsmath,amsfonts,bm}

\def\eqref#1{equation~\ref{#1}}

\def\1{\bm{1}}

\DeclareMathAlphabet{\mathsfit}{\encodingdefault}{\sfdefault}{m}{sl}
\SetMathAlphabet{\mathsfit}{bold}{\encodingdefault}{\sfdefault}{bx}{n}

\newcommand{\softmax}{\mathrm{softmax}}

